\documentclass[twoside]{article}
\usepackage[dvipsnames]{xcolor}
\usepackage{graphicx}
\usepackage{amsthm}
\usepackage{amsmath}
\usepackage{mathtools}
\usepackage{mathrsfs}
\usepackage{amssymb}
\usepackage{lipsum}
\usepackage{etoolbox}
\usepackage[linesnumbered,lined,boxed,ruled]{algorithm2e}
\usepackage[section]{placeins}
\usepackage[breaklinks,colorlinks,citecolor=blue,linkcolor=blue,urlcolor=blue]{hyperref}
\usepackage[all]{hypcap}
\usepackage{cancel}
\usepackage{csquotes}
\usepackage[toc,page]{appendix}
\usepackage{multirow}
\usepackage{datetime}
\usepackage{times}
\usepackage[format=plain, labelfont={bf,it}, textfont=it]{caption}
\usepackage{tcolorbox}
\usepackage{natbib, bibentry}
\usepackage{orcidlink}
\usepackage{fancyhdr}

\usepackage{scalerel}
\usepackage{stackengine,wasysym}

\newdateformat{mdydate}{\monthname[\THEMONTH] \THEDAY, \THEYEAR}
\newdateformat{monthyeardate}{\monthname[\THEMONTH] \THEYEAR}
\newdateformat{vmonthyeardate}{\THEYEAR.\THEMONTH.\THEDAY}

\newcommand{\be}{\begin{equation}}
\newcommand{\ee}{\end{equation}}
\newcommand{\bea}{\begin{eqnarray}}
\newcommand{\eea}{\end{eqnarray}}

\usepackage{geometry}
\graphicspath{{figures}}

\newcommand{\volume}{2}
\newcommand{\firstpage}{40}
\newcommand{\lastpage}{56}
\newcommand{\yyyy}{2025}
\newcommand{\mm}{October}
\newcommand{\dd}{9}
\newcommand{\authors}{Pfeffer et al.}
\newcommand{\fulltitle}{Context-Aware Inference via Performance Forecasting\\in Decentralized Learning Networks}
\newcommand{\shorttitle}{Context-Aware Inference via Performance Forecasting}

\newcommand{\doi}{10.70235/allora.0x\volume\ifnum\numexpr\firstpage<10 000\else\ifnum\numexpr\firstpage<100 00\else\ifnum\numexpr\firstpage<1000 0\fi\fi\fi\firstpage}
\fancypagestyle{firstpage}{%
    \fancyhead[L]{\footnotesize Allora Decentralized Intelligence \textbf{\volume}, \firstpage--\lastpage; \yyyy\ \mm\ \dd}
    \fancyhead[R]{\footnotesize doi:\href{https://doi.org/\doi}{\textcolor{blue}{\doi}}}
    \fancyfoot{}
    
}

\AtBeginDocument{\thispagestyle{firstpage}}

\begin{document}

\title{\fulltitle}
\author{\authors}
\date{\monthyeardate{\today}}


\vskip30mm
\begin{center}
\begin{minipage}{170mm}
\begin{center}
\vskip5mm
{\fontsize{15pt}{15pt}\textbf{Context-Aware Inference via Performance Forecasting\\in Decentralized Learning Networks}}
\vskip5mm
Joel Pfeffer$^{\orcidlink{0000-0003-3786-8818}}$$^{1}$,
J.~M.~Diederik Kruijssen$^{\orcidlink{0000-0002-8804-0212}}$$^{1}$,
Cl\'{e}ment Gossart$^{\orcidlink{0009-0006-6509-6041}}$$^{1}$,
M\'{e}lanie Chevance$^{\orcidlink{0000-0002-5635-5180}}$$^{1,2}$,\\
Diego Campo Millan$^{\orcidlink{0009-0000-0866-3691}}$$^{1}$,
Florian Stecker$^{\orcidlink{0000-0002-7687-5116}}$$^{1}$ \&
Steven~N.~Longmore$^{\orcidlink{0000-0001-6353-0170}}$$^{1,3}$
\vskip1mm
$^{1}$\textit{Allora Foundation},
$^{2}$\textit{Cosmic Origins Of Life (COOL) Research DAO},
$^{3}$\textit{Liverpool John Moores University}
\end{center}
\end{minipage}
\end{center}
\vspace{3mm}

\begin{abstract}
\noindent
In decentralized learning networks, predictions from many participants are combined to generate a network inference. While many studies have demonstrated performance benefits of combining multiple model predictions, existing strategies using linear pooling methods (ranging from simple averaging to dynamic weight updates) face a key limitation. Dynamic prediction combinations that rely on historical performance to update weights are necessarily reactive. Due to the need to average over a reasonable number of epochs (e.g.\ with moving averages or exponential weighting), they tend to be slow to adjust to changing circumstances (e.g.\ phase or regime changes). In this work, we develop a model that uses machine learning to forecast the performance of predictions by models at each epoch in a time series. This enables `context-awareness' by assigning higher weight to models that are likely to be more accurate at a given time. We show that adding a performance forecasting worker in a decentralized learning network, following a design similar to the Allora network, can improve the accuracy of network inferences. Specifically, we find that forecasting models that predict regret (performance relative to the network inference) or regret $z$-score (performance relative to other workers) show greater improvement than models predicting losses, which often do not outperform the naive network inference (historically weighted average of all inferences). Through a series of optimization tests, we show that the performance of the forecasting model can be sensitive to choices in the feature set and number of training epochs. These properties may depend on the exact problem and should be tailored to each domain. Although initially designed for a decentralized learning network, using performance forecasting for prediction combination may be useful in any situation where predictive rather than reactive model weighting is needed.
\end{abstract}
\vspace{3mm}


\section{Introduction}
\label{sec:intro}

Models are naturally an approximation of real processes and may have their own unique data sources (private information), assumptions, uncertainties and biases. Some models may also perform better than others in certain contexts, but worse at other times, meaning there is generally no single best model that outperforms the others under all conditions. For this reason, combining the predictions from multiple models (often termed forecast combination or aggregation) can lead to substantial improvements in accuracy over the individual models.

Beginning with \citet{Reid_68} and \citet{Bates_and_Granger_69}, numerous methods have been developed for model aggregation \citep[for reviews, see][]{Clemen_89, Timmermann_06, Wang_et_al_2023}. The aim is to identify the optimal set of weights to combine the predictions, such that the error of the resulting combined prediction is minimized. The success of combination techniques naturally depends on the accuracy and diversity of the underlying models. If the set of models is carefully chosen (i.e.\ to reduce bias and remove poorly performing models) then simple averaging strategies often perform well \citep[mean, median, trimmed mean, etc.; e.g.][]{Clemen_and_Winkler_86, Batchelor_and_Dua_95, Stock_and_Watson_04, Genre_et_al_13, Lichtendahl_and_Winkler_20, Petropoulos_and_Svetunkov_20}. A natural extension to such methods is to use linear combinations with non-equal weights \citep[e.g.][]{Newbold_and_Granger_74, Granger_and_Ramanathan_84, Kolassa_11}. Such strategies have been extended by using machine learning (often termed meta-learning) to assign combination weights \citep[e.g.][]{Prudencio_and_Ludermir_06, Lemke_and_Gabrys_10, Montero-Manso_et_al_20, Kang_et_al_22} or select the best model for a particular time series \citep[e.g.][]{Kuck_et_al_16, Gastinger_et_al_21, Talagala_et_al_23}.

Of course, simple averaging does not take into account potential changes in relative performance over time (e.g.\ due to time-varying trends, seasonality changes or structural breaks), leading to the development of time-varying combination methods. Time-varying approaches generally update the combination weights using recent historical data (e.g.\ with a rolling window), such as through exponential weighting or moving averages \citep[e.g.][]{Bates_and_Granger_69, Diebold_and_Pauly_87, Aiolfi_and_Timmermann_06}, parametric methods with smooth transitions or switching \citep[e.g.][]{Sessions_and_Chatterjee_89, LeSage_and_Magura_92, Deutsch_et_al_94, Elliott_and_Timmermann_05}, and non-parametric methods that do not impose functional forms on the coefficients \citep[e.g.][]{Terui_and_van_Dijk_02, Chen_and_Maung_23}. Other methods involve learning weights using data restricted to be similar to the current time \citep[such as from certain phases in periodic data,][]{Dudek_25}. Recent work has also used neural networks to combine predictions at each epoch in a time series, without estimating weights \citep{Zhao_and_Feng_20}.

Artificial intelligence, including machine learning, has revolutionized many fields. In the case of time-series predictions, machine learning enables predictions to be made using data-driven approaches without detailed knowledge of the underlying processes, and now dominates forecasting competitions \citep{Makridakis_et_al_20, Makridakis_et_al_22}. With the availability of numerous high-performing open source algorithms \citep[e.g.][]{Chen_and_Guestrin_16, Ke_et_al_17, Oreshkin_et_al_20} and ever-increasing computing power, the barrier-to-entry is lower than ever.

However, even with such improvements, increasing machine learning performance requires increased computational resources \citep{Makridakis_et_al_20}. Building on decades of research into model combinations, decentralized learning networks address this issue by enabling participants to collaborate under complete data and model privacy by submitting their own inferences or predictions to a network that generates a combined `network inference' for the target variable \citep{Craib_et_al_17, Rao_et_al_21, Steeves_et_al_22, K24}. This allows diverse sets of models to be combined independently of their composition (e.g.\ with differing algorithms, features and private data sets), a key requirement of model combination strategies \citep{Bates_and_Granger_69, Batchelor_and_Dua_95, Lichtendahl_and_Winkler_20, Kang_et_al_22}. Of course, the question remains what the optimal method is for combining all predictions, and potential issues with model combination are exacerbated in decentralized learning networks: there is no guarantee that predictions are diverse and unbiased due to its decentralized nature; the set of models may change over time (e.g.\ due to participants joining and leaving the network, or through the dynamic selection of subsets of participants, see \citealt{K24_sortition}).
In such a case, simple averages of predictions will perform poorly, and dynamic weighting is required as the network evolves over time.

A number of strategies have been developed to weight predictions in decentralized networks, such as stake-weighted averaging \citep{Craib_et_al_17} and peer-ranked weights based on historical performance \citep{Rao_et_al_21, Steeves_et_al_22}. However, such methods cannot identify which models are likely to be most accurate in different conditions (e.g.\ seasonality changes or structural breaks). The Allora network \citep{K24} solves this problem by introducing workers that \textit{forecast} the expected performance of worker inferences at each epoch. The aim of the forecasting workers is not to determine a constant or slowly varying set of weights for each worker (e.g.\ based on historical performance), but to learn under which conditions each worker may be more accurate (i.e.\ context awareness). Given the non-uniqueness of weights at a single epoch and the time-evolution of the participant set (implying that the optimal weights will change), the forecasting workers predict performance (losses, regrets, or regret $z$-scores) rather than weights. The predicted losses are then transformed into weights by the network through an optimized sigmoid function (often referred to as a `scaled logistic gate') to create `forecast-implied' inferences \citep{K25_optimizing}.

In this work, we describe a forecasting model designed for the Allora network, although the principles could be extended to any model combination problem where predictive rather than reactive weighting is important. The paper is structured as follows. In \S\ref{sec:design}, we describe the relevant structure of the decentralized learning network and design of the forecasting model, including machine learning models, target variables and feature sets. In \S\ref{sec:benchmarks}, we test the forecaster model against a series of synthetic benchmarks, aiming to optimize the model for context awareness. We then extend the benchmark tests to experiments with real data from the Allora testnet in \S\ref{sec:experiments}. Finally, in \S\ref{sec:conclusions} we discuss the findings and summarize the conclusions of the paper.

\section{Forecaster design}
\label{sec:design}
The goal of this work is to design a forecasting model that functions within a decentralized learning network and can identify when inferences from each participant are likely to be more accurate than others, i.e.\ is `context aware', such that it can outperform a `naive' network inference (a historically weighted average of all inferences). There are a number of critical elements to consider in the design of such a forecaster model, such as the underlying machine learning model, the forecasted variable, and the feature set. These elements depend on the learning network design and the target variable.

\subsection{Decentralized learning network}
\label{sec:network}
We consider a decentralized learning network based on the Allora network \citep{K24}. Allora consists of `topics', which are sub-networks within which the participants collaborate to achieve a single common objective. A topic consists of $N_\mathrm{i}$ `inference workers' providing inferences and $N_\mathrm{f}$ `forecasting workers' providing forecasts for the performance of the inference workers. At each epoch $i \in \{1, \dots, N_\mathrm{e} \}$, a network inference $I_i$ is generated as follows.

Each inference worker $j \in \{1, \dots, N_\mathrm{i} \}$ submits an inference
\begin{equation}
    I_{ij} = M_{ij} ( D_{ij} ) ,
\end{equation}
using its own dataset $D_{ij}$ and model $M_{ij}$.
Once the true value becomes available, the true inference loss $L_{ij}$ and instantaneous regret $R_{ij} = L_i - L_{ij}$ are evaluated against the ground truth using the loss function for the topic, where $L_i$ is the true loss of the network inference.

For each inference $I_{ij}$ submitted by inference worker $j$, each forecasting worker $k \in \{1, \dots, N_\mathrm{f} \}$ submits forecasted losses
\begin{equation}
    \log L_{ijk} = M_{ijk} ( D_{ijk} ) ,
\end{equation}
using its own dataset $D_{ijk}$ and model $M_{ijk}$. For each forecasting worker $k$, the raw inferences $I_{ij}$ are combined with the forecasted losses $L_{ijk}$ to generate a forecast-implied inference
\begin{equation}
    I_{ik} = \frac{\sum_j w_{ijk} I_{ij}}{\sum_j w_{ijk}},
\end{equation}
where the weights are a function of the forecasted losses $w_{ijk}(L_{ijk})$.

To calculate the weights $w_{ijk}$, the forecasted losses are first converted to approximate forecasted regrets
\begin{equation} \label{eq:regret}
    R_{ijk} = \log \mathcal{L}_{i-1} - \log L_{ijk} ,
\end{equation}
using the network loss at the previous epoch $\mathcal{L}_{i-1}$ (this approximation is necessary, because the network loss of the current epoch is not available yet). Next, to ensure consistency from epoch to epoch, the forecasted regrets are converted to normalized regrets as
\begin{equation} \label{eq:norm_regret}
    \hat{R}_{ijk} = \frac{R_{ijk}}{\sigma_j(R_{ijk}) + \epsilon} ,
\end{equation}
where $\sigma_j(R_{ijk})$ is the standard deviation of forecasted regrets and $\epsilon$ is a small value added to avoid division by zero.

Finally, the approximate regrets are converted to weights as
\begin{equation} \label{eq:weights}
    w_{ijk} = \phi'(\hat{R}_{ijk}) ,
\end{equation}
using a sigmoid weighting function
\begin{equation} \label{eq:weight_fn}
    \phi'(x) = \frac{p}{e^{-p(x - c)} + 1} ,
\end{equation}
with fiducial values $p=3$ and $c = 0.75$, where $c$ is the transition point between positive contributions with a constant non-zero weight and negative contributions with a scaling $w_{ijk} \propto L_{ijk}^{-p}$. In this way, inferences that are forecasted to be more accurate than the network at the previous epoch are given more weight than those forecasted to be less accurate.

The inferences (both raw and forecast-implied) from all workers $l \in \{1, \dots, N_\mathrm{i} + N_\mathrm{f} \}$ are combined to generate the network inference
\begin{equation} \label{eq:network_inference}
    I_{i} = \frac{\sum_l w_{il} I_{il}}{\sum_l w_{il}} ,
\end{equation}
where the weights $w_{il} = \phi'(\mathcal{R}_{i-1,l})$ are set by an exponential moving average (EMA) of each worker's actual regret at the previous epoch $\mathcal{R}_{i-1,l}$ \citep[the EMA uses a fiducial value $\alpha = 0.1$,][]{K25_optimizing}. For reference, a `naive' network inference $I_{i}^{-}$ is also calculated using only the raw inferences (i.e.\ omitting the forecast-implied inferences from \autoref{eq:network_inference}).

In practice, to improve network performance (both in terms of accuracy and computational cost), the set of inferences combined to generate network inferences may be limited to an `active set' (typically 32 inferers on Allora) through merit-based sortition \citep{K24_sortition}. In that case, forecasts are only generated for the active set of inferers.

\subsection{Forecaster model}
\label{sec:model}
We use the gradient-boosted decision tree models XGBoost \citep{Chen_and_Guestrin_16} and LightGBM \citep{Ke_et_al_17} as a machine-learning basis for the forecaster model. For tabular data (samples with the same set of features), decision tree models typically outperform and have shorter training times than deep neural networks \citep[e.g.][]{Shwartz-Ziv_and_Armon_22, Borisov_et_al_24}. For both XGBoost and LightGBM, we perform automated hyperparameter optimization with Optuna \citep{Akiba_et_al_19}.

We consider two main structures for the forecaster model: whether to train a single global forecasting model (with inferer ID as a feature to enable context awareness), or whether to train independent models for each inferer. Both methods have their own strengths and weaknesses. A global model may be able to combine information from similar workers to enable stronger predictions from an increased sample size, but may also be susceptible to training on population-averaged performance, rather than identifying context from individual workers. Conversely, a series of per-inferer models prevents crossover of information between different workers, but must train on smaller amounts of data and has a higher overhead (e.g.\ increased training times). To maximize context awareness, the fiducial setup of the forecasting model uses a series of per-inferer models trained on the active set of inferers, but uses a global forecasting model as a fallback for inferers yet to be trained (those with insufficient data, such as new inferers).

\subsubsection{Forecaster target variable}
The Allora network was originally designed such that forecasting models predict the losses for each inferer, which are then used to define the weights of a `forecast-implied inference' that combines the raw inferences through a weighted sum. In the experiments carried out for this study, we find that different forecaster target variables can be optimal in different circumstances.
Therefore, we consider three main forecaster targets, each with their own strengths and weaknesses:
\begin{description}
    \item[Loss] Forecasting losses has the benefit that the loss of each inference is independent; it is simply a measure of the accuracy of an inference. However, losses can be challenging to predict if there is high variability. As losses require conversion to regrets for the weighting function, and regrets must be approximated using the network loss at the previous epoch (\autoref{eq:regret}), significant epoch-to-epoch changes in the network loss can also degrade forecaster performance if the weighting function is non-linear with regret (the Allora network adopts a sigmoid form).
    \item[Regret] Forecasting the regrets of inferences bypasses the loss-to-regret conversion (\autoref{eq:regret}) and potentially provides a more stable property to predict: it is only performance relative to the network that must be forecasted, rather than the absolute performance of each inferer (loss). A potential disadvantage is that regret can depend upon the makeup of the network: if the set of inferers at each epoch changes significantly over time, the regret of an inferer could change without a corresponding change in inference accuracy.
    \item[Regret $z$-score] Rather than forecasting performance relative to the network inference (regret), the regret $z$-score considers performance only relative to other inference workers.
    Similar to normalized regrets (\autoref{eq:norm_regret}), we define the regret $z$-score for each inference as
    \begin{equation} \label{eq:zscore}
        Z_{ij} = \frac{R_{ij} - \left< R_{ij} \right> }{\sigma_j(R_{ij}) + \epsilon} ,
    \end{equation}
    where $\left< R_{ij} \right>$ and $\sigma_j(R_{ij})$ are the mean and standard deviation of all inferences at epoch $i$, and $\epsilon$ is a small value to avoid division by zero. Note that the network loss cancels out, so that $Z_{ij}$ is independent of $\mathcal{L}_{i-1}$. The regret $z$-score then replaces the normalized regret in the weighting function: $w_{ijk} = \phi'(Z_{ijk} + \delta_Z)$, with a fiducial offset $\delta_Z = -1$ to shift the transition point in the weight function (\autoref{eq:weight_fn}), which was found to improve performance. As with forecasting regrets, the regret $z$-score may be affected by changes in inferer composition, particularly in cases of small numbers of inferers where the standard deviation of regret is poorly defined. As regret is simply negative log loss with an offset, the $z$-score target could be equivalently written in terms of negative losses.
\end{description}
These considerations suggest that, rather than a single best model, a suite of forecasters with different target variables may be required to adapt to different situations. 

\subsubsection{Feature set}
We consider two types of feature data that the models will use to make predictions. First, we construct a baseline dataset that contains properties from the network (e.g.\ inferer performance metrics such as losses and regrets). Secondly, we assume the existence of a `private' dataset that contains additional information compiled by the forecaster model builder. This data set is domain-specific (e.g.\ market prices for the particular topic).
\begin{description}
    \item[Baseline data] The complete set of baseline properties are: inferer ID, worker inference values (current epoch), worker losses and regrets (previous epoch), network loss (previous epoch), worker rewards and performance score (previous epoch; see \S4.1 of \citealt{K24}). This set is then extended through feature engineering to capture the key dynamics of these properties. We apply the following transformations to the worker inferences, losses and regrets: gradient, momentum, acceleration, exponentially-weighted mean and standard deviation, rolling mean and standard deviation, difference from moving average, and autocorrelation (see below). We also consider the mean, standard deviation and $z$-score of the losses and regrets of all workers at each epoch. For the exponentially-weighted and rolling properties, we consider the epoch span as a variable to be optimized for the particular topic, but generally find short spans provide best performance (a combination of 3 and 14 epochs are used by default).
    \item[Private data] The set of private features depends on the particular property that is being predicted, but should generally benefit from being similar to the features used by workers to generate inferences. For simplicity, our tests consider network topics that predict prices in financial markets (though this is by no means a restriction). In this case, the base set of private features should be the historical properties of an asset (e.g.\ open, high, low, close prices and volume). As above, this set can be extended through feature engineering to capture key market dynamics, e.g.\ moving and exponentially weighted averages, momentum, percentage change, volatility, Bollinger Bands, price ratios. Many of these quantities generalize well to any other online target variable.    
\end{description}

We use autocorrelation (correlation of values in a time series separated by a given lag) as part of feature engineering to identify any regular periodic signals in the feature data. Where correlations at given lags are significant, features shifted by those lags are added to the feature set (for lags $\geq 2$ epochs, since features at the previous epoch are already included). In practice, we use both the autocorrelation function (ACF) and partial autocorrelation function (PACF) to identify significant lags. The PACF controls for correlations due to multiples of shorter lags (e.g.\ factors of two) and we find it can identify lags that are not significant in the standard ACF. Therefore, we only include lags which have a significance $>99\%$ in both ACF and PACF.

A large number of feature variables can reduce the performance of a model by increasing training and prediction times due to increased complexity, and increasing the risk of overfitting, such that the model cannot generalize. First, we remove any features with zero variance. Next, we identify pairs of highly correlated features (with Pearson product-moment correlation coefficients $>0.95$) and keep only the feature that has the strongest correlation with the forecaster target variable. These steps typically reduce the feature set from $\approx 80$ to $\approx 50$ features. We experimented with other forms of automated feature set reduction (e.g.\ recursive feature elimination, Boruta selection, variance thresholds), but found they did not adequately identify features with the highest importance in the machine learning models.

We use 1000 epochs as the default number of training epochs. We find that this provides a reasonable balance between training time, having sufficient data for training, and focusing on recent data. Using a much larger number of training epochs can degrade forecaster performance if the earliest information is no longer relevant for the current inferer models (e.g.\ if inferer models have since improved or worsened).

\section{Synthetic benchmarks}
\label{sec:benchmarks}
We begin testing the forecaster models with a series of synthetic benchmarks of varying levels of complexity. The aim is to use the tests to identify the best performing forecaster model(s) (target variable, global/per-inferer model) and the optimal set of features. In these benchmarks, the forecasters do not contribute to the combined network inferences so that a direct comparison can be made between different forecasting models. We show results using the LightGBM machine learning model, but the conclusions are unaltered when comparing results from XGBoost. For all tests we use a mean-squared error (MSE) loss function.

\subsection{Periodic outperformance}
\label{sec:periodic_benchmark}
\begin{figure}[!]
    \centering
    \includegraphics[width=0.495\linewidth]{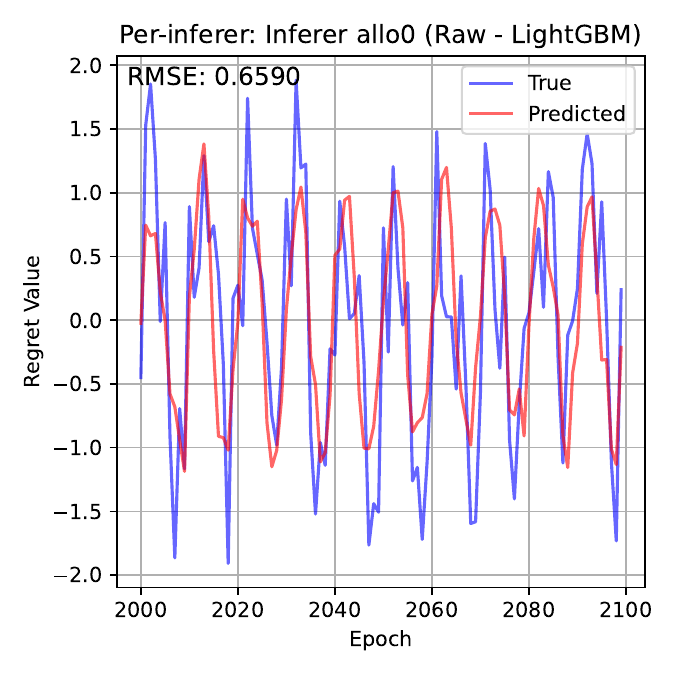}
    \includegraphics[width=0.495\linewidth]{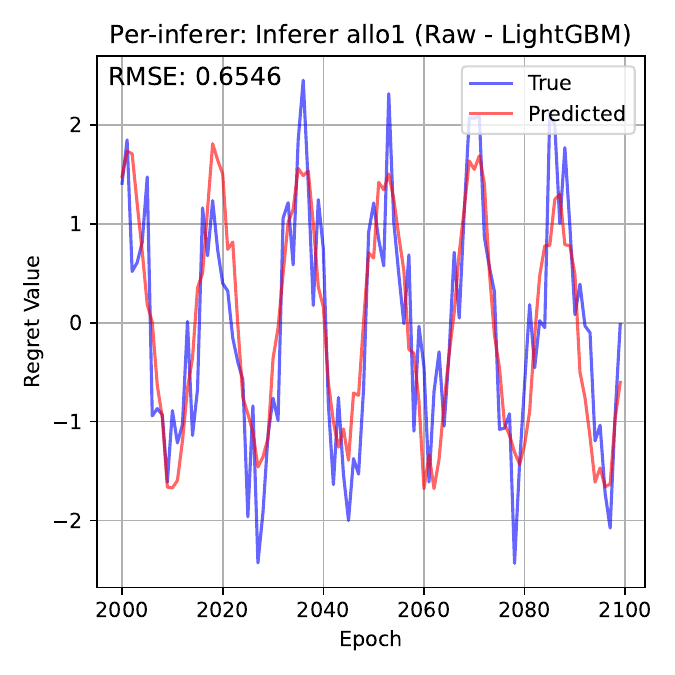}
    \caption{True and predicted regret values for two `inferers' in a simple periodic (sinusoidal) outperformance test. In the left panel the sine function has an amplitude of 1 and period of 10 epochs, while in the right panel the sine function has an amplitude of 1.5 and period of 17 epochs. A random noise term (uniformly sampled $\in[-1,1]$) is added at each epoch for both inferers. In both cases, the default per-inferer forecasting model reasonably identifies both the amplitudes and periods of the underlying sinusoidal evolution for each worker.}
    \label{fig:sine_evo}
\end{figure}
In the first test, the regrets for two inferers follow a simple sinusoidal evolution with an additional random component (uniform random errors in the range $\pm 1$). The sine function for each inferer has a different amplitude (1 and 1.5) and period (10 and 17 epochs). Eight other inferers with only random regrets (in the range $\pm 1$) are added as a baseline. The test period is 100 epochs with 1000 epochs used for training. Though rather unrealistic, this test helps to understand the behaviour in predictions from different forecasting models.

The evolution of the true and predicted regrets for the two periodically outperforming workers (\textit{allo0} and \textit{allo1}) are shown in \autoref{fig:sine_evo} for the default per-inferer forecaster model. In the models for both outperforming inferers, autocorrelations of the regrets tend to be the most important features, along with other relative change variables (momentum, acceleration, percentage change). For this test, models without autocorrelation show reduced performance, with the root mean square error (RMSE) of the models increasing to $0.709$ and $0.669$ (compared to $0.659$ and $0.655$ for inferers \textit{allo0} and \textit{allo1}, respectively). This is due to the model undershooting the peaks of the sine function (predicting $\approx 0.7$ instead of $\approx 1$), but otherwise it reasonably predicts the sinusoidal evolution through the value at the previous epoch, exponentially-weighted and rolling mean, gradient and percentage change.

\begin{figure}
    \centering
    \includegraphics[width=0.495\linewidth]{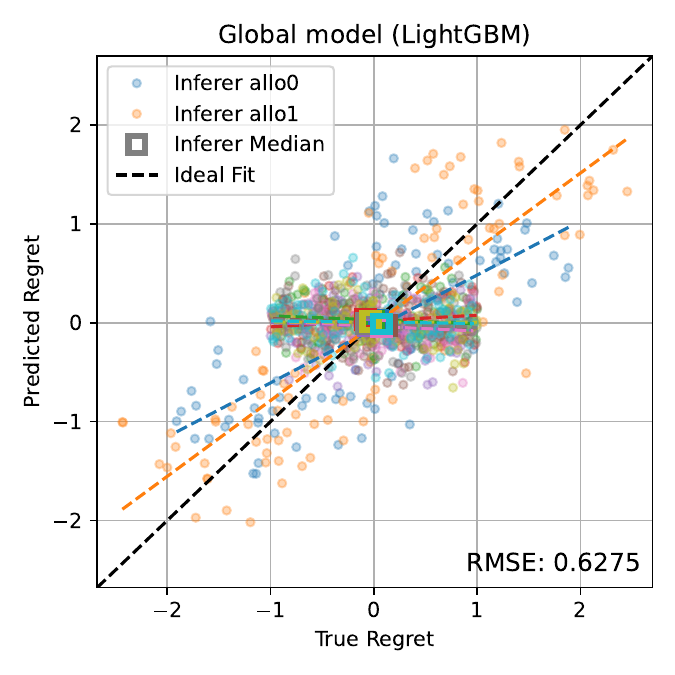}
    \includegraphics[width=0.495\linewidth]{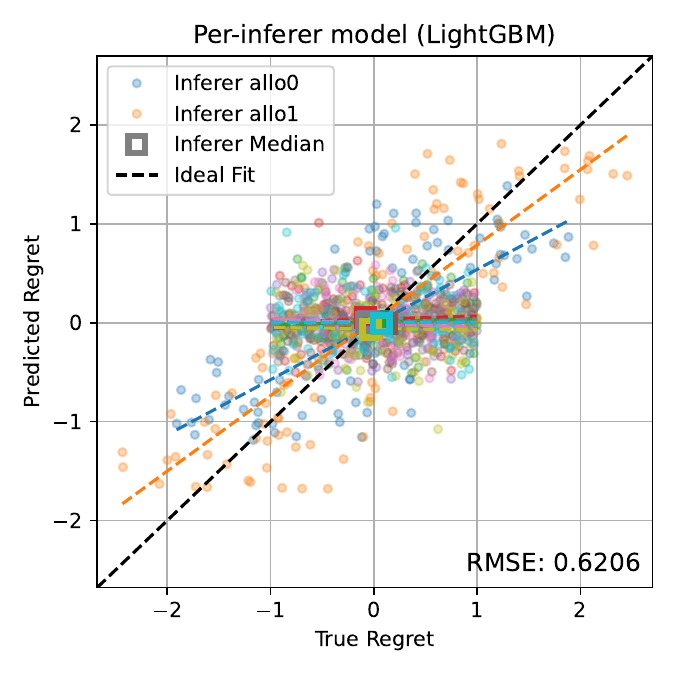}
    \caption{Comparison of true and predicted regrets in the sinusoidal evolution test for the global (left panel) and per-inferer (right panel) forecasting models. Small data points show individual values for each inferer at each epoch and large open squares show the median of true and predicted regrets for each inferer. Coloured dashed lines show linear fits for each inferer, using Huber regression to minimize effects of outliers. The black dashed line shows the ideal one-to-one relation. Though the overall performance is similar (as indicated by the RMSE in each panel), the global model shows some confusion in predicted regrets between the two `outperforming' workers (\textit{allo0} and \textit{allo1}, e.g.\ at values $< 0$), while the per-inferer model clearly distinguishes their predictions (clear bands at $-1$ and $-1.5$, respectively).}
    \label{fig:sine_pred}
\end{figure}
In \autoref{fig:sine_pred}, we compare the true and predicted regrets for all inferers for the global forecasting model (left panel) and the per-inferer model (right panel). We consider this comparison as the `context awareness' test, i.e.\ it tests the ability of the forecasting model to predict outperformance or underperformance by the inferers. In this test, the global model and the per-inferer model have very similar overall performance ($\mathrm{RMSE} = 0.628$ and $0.621$, respectively). However, the per-inferer model performs better for the outperforming inferers ($\mathrm{RMSE} = 0.659$ and $0.655$ for inferers \textit{allo0} and \textit{allo1}, respectively) than the global model ($\mathrm{RMSE} = 0.709$ and $0.669$ for inferers \textit{allo0} and \textit{allo1}, respectively), which can also be seen in the increased gradients for the linear fits in the figure. The difference in the overall RMSE is due to the decreased scatter for the random inferers in the global model, for which the optimum strategy is simply to predict the mean. This indicates that the per-inferer model is better at distinguishing workers, which can (for example) be seen in \autoref{fig:sine_pred} by comparing the predicted regrets at values $< 0$ for inferers \textit{allo0} and \textit{allo1}: the per-inferer model shows clear banding for both workers, while the global model shows more confusion between the predictions for the workers. In contrast, the global model can benefit from an increase in the training data set by stacking results for similarly-performing inferers.

\begin{figure}
    \centering
    \includegraphics[width=0.495\linewidth]{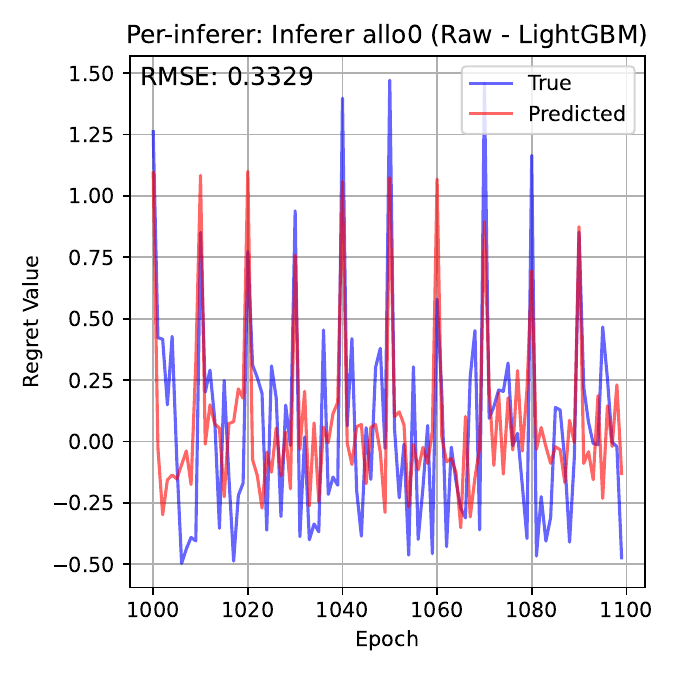}
    \includegraphics[width=0.495\linewidth]{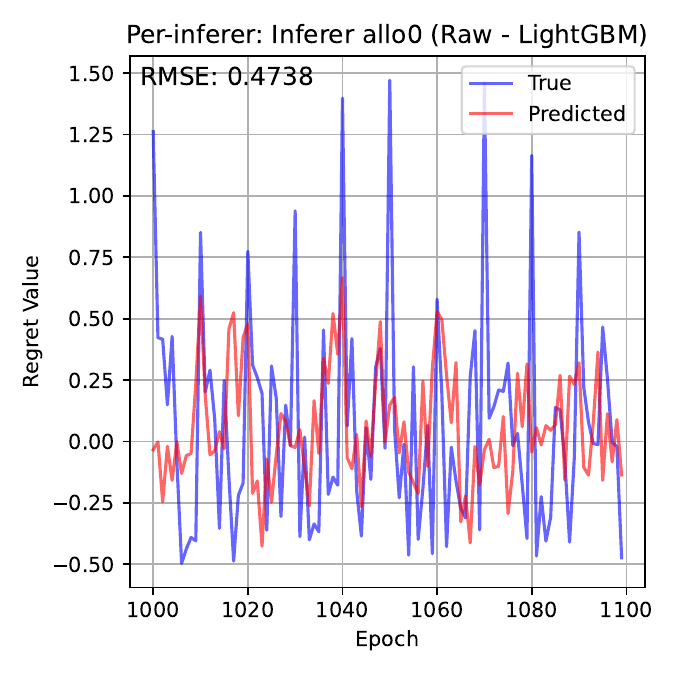}
    \caption{Fixed interval periodic outperformance test ($+1$ in regret every 10 epochs), with (left) and without (right) autocorrelation for the per-inferer forecasting model. This test shows that autocorrelation is crucial to identify the outperformance epochs.}
    \label{fig:periodic_epoch}
\end{figure}
As an extension to the sinusoidal regrets test, in \autoref{fig:periodic_epoch} we compare results for an inferer with outperformance at fixed intervals ($+1$ in regret every 10 epochs) and otherwise random performance (uniform random regrets between $-0.5$ and $0.5$). This test models an inferer that outperforms at regular times (e.g.\ time of day or day of the week). The figure compares results for a forecasting model with and without autocorrelation (left and right panels, respectively). Naturally, in this test the model with autocorrelation ($\mathrm{RMSE} = 0.333$) significantly outperforms the model without autocorrelation ($\mathrm{RMSE} = 0.474$). For the model without autocorrelation, the most important features are exponentially-weighted and rolling means of the regret, which fail to sufficiently identify the outperformance epochs.

As for the sinusoidal test, when a second outperforming inferer is added with relatively similar outperformance peaks ($+1.25$ in regret every 17 epochs), the global forecasting model does not distinguish the amplitude of the outperformance peaks between the two inferers, predicting $+1$ in both cases. The global model generally distinguishes the differing periods of the workers, but occasionally misses outperformance epochs (particularly the second inferer with a period of 17 epochs). By design, the per-inferer model cannot confuse the amplitudes or periods for either worker.

\subsection{Contextual outperformance}
\label{sec:contextual_benchmark}
As a more realistic test, we now consider an experiment that uses geometric brownian motion to generate true values, with an initial value of $1000$ and volatility of $0.01$. The drift parameter is randomly modulated between values of $[-0.01, 0, 0.01]$ (downward drift, no drift and upward drift, respectively) for periods with a typical length of 5 epochs (with each period length drawn from a Poisson distribution with an expectation value of 5). The zero drift periods have a probability of occurrence that is three times higher than the drift periods. As in \S\ref{sec:periodic_benchmark}, we use $1000$ epochs for training and $100$ epochs for testing.

\begin{figure}[!t]
    \centering
    \includegraphics[width=0.95\linewidth]{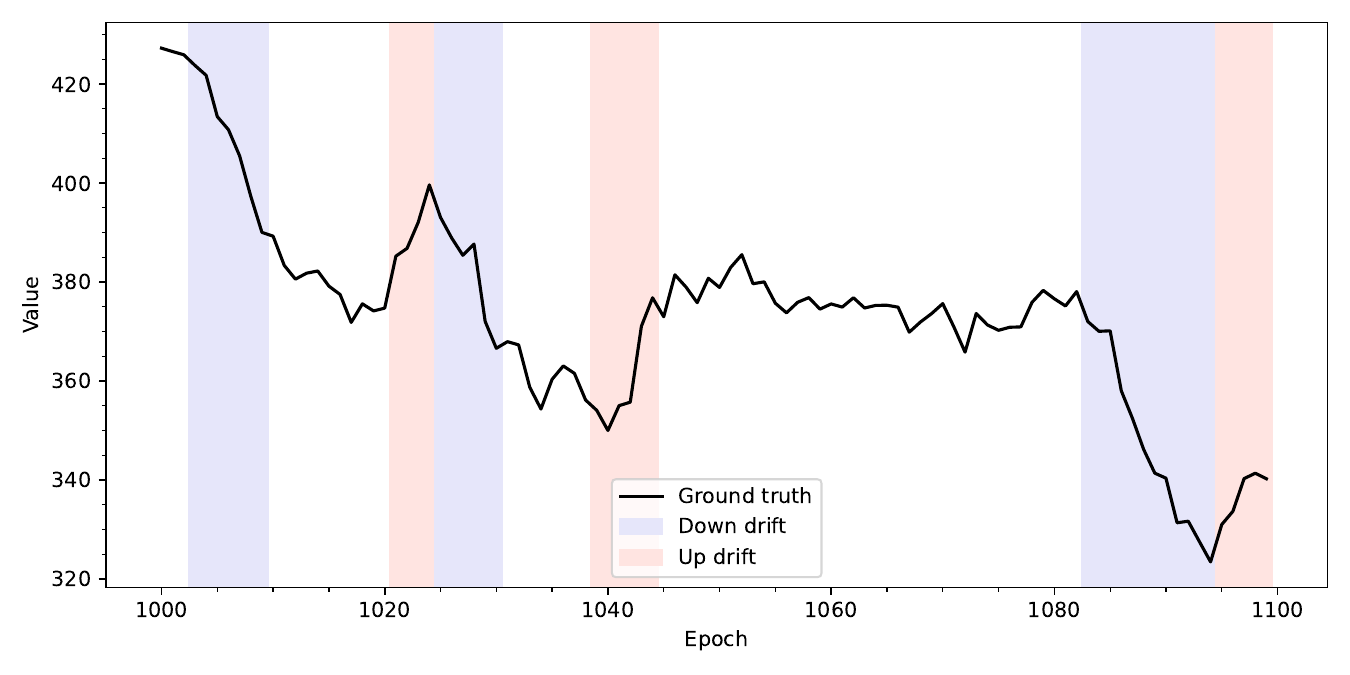}
    \includegraphics[width=0.95\linewidth]{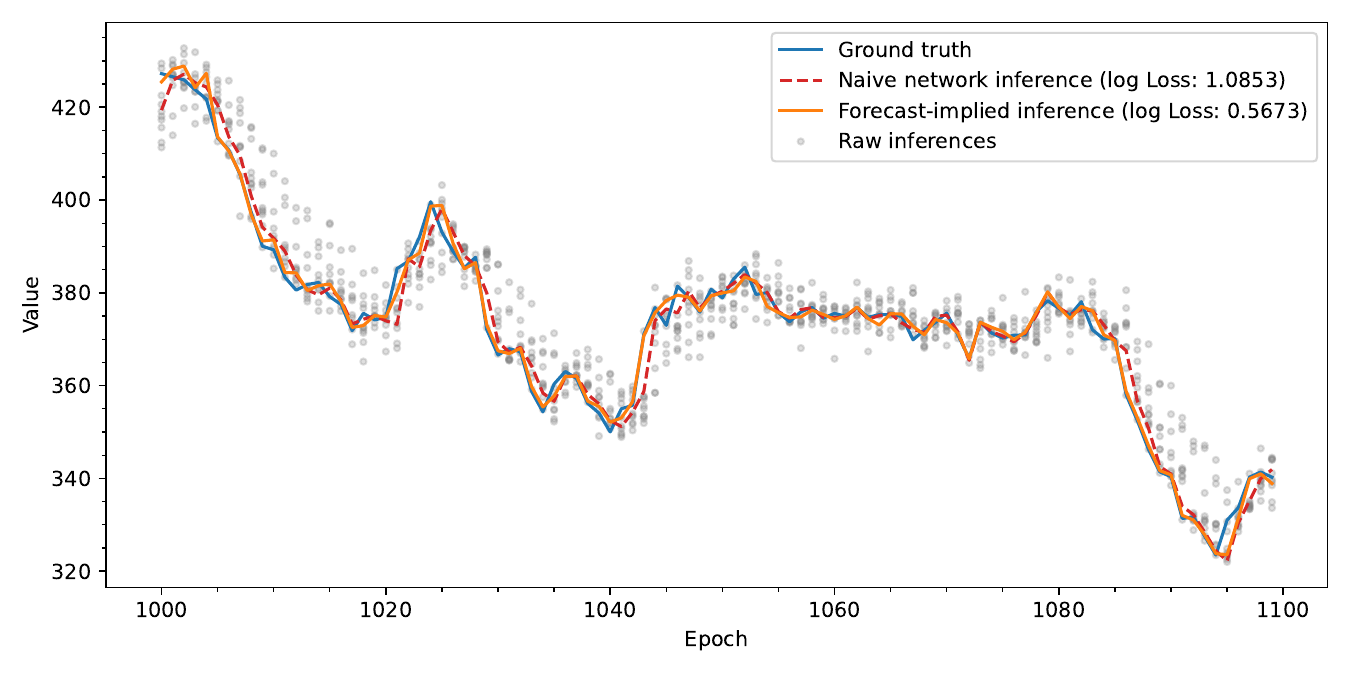}
    \caption{Top: Ground truth for the contextual outperformance test, with shaded regions highlighting the periods with a non-zero drift parameter (blue: downward drift; red: upward drift). Bottom: Comparison of individual inferences (grey points), naive network inference (red dashed line), and forecast-implied inference (solid orange line) with the ground truth (solid blue line). The forecast-implied inference was generated using forecasted $z$-scores from the default per-inferer forecaster model. The mean log loss of the naive network and forecast-implied inferences are indicated in the legend, with the forecast-implied inference ($\log{L}=0.57$) clearly outperforming the naive network inference ($\log{L}=1.09$).}
    \label{fig:context_drifts}
\end{figure}
The top panel of \autoref{fig:context_drifts} shows the evolution of the ground truth generated for the contextual outperformance test. The shaded regions highlight the periods with a non-zero drift parameter (where blue indicates downward drift and red indicates upward drift). As shown in the figure, periods of some drifting can occur even with a drift parameter of zero (at an epoch $\approx 1035$), but they tend to have shorter durations than the controlled drift periods.

We then create ten inferers, three of which outperform in different circumstances (\textit{allo0}: downtrends, \textit{allo1}: uptrends, \textit{allo2}: no drift), along with seven control inferers which predict random log returns (i.e.\ the natural logarithm of the ratio of the new price to the previous price). We choose to generate the predictions in log returns due to its compounding effect across a time series. Each of the inferers have random log returns drawn from a Gaussian distribution with the standard deviation set by the volatility of the true values scaled by a random factor depending on their outperformance. During their accurate periods, the three outperforming inferers use a random factor in the range $0.1$--$0.3$, with the random returns added to the true returns, and otherwise predict only random returns with a random volatility factor in the range $0.5$--$1$. Four random inferers predict only random returns at all times, with a random volatility factor in the range $0.2$-$1.2$. The final three random inferers use a random volatility factor in the range $0.5$-$1$ to predict returns relative to an EMA of the ground truth (with spans $5$, $7$ and $9$). This behaviour was chosen to drive delayed reactions during drift periods, such that they become \textit{under}-performing workers.

The bottom panel of \autoref{fig:context_drifts} compares the ground truth values (solid blue line) with the individual inferences (grey points), naive network inference (dashed red line) and forecast-implied inference from the default per-inferer model predicting regret $z$-scores (solid orange line). For this comparison, all inferences were transformed from log returns back to absolute price space. In this test, the forecast-implied inference (mean log loss $\log{L}=0.57$) clearly outperforms the naive network inference ($\log{L}=1.09$). This is most noticeable during the controlled drift periods, where the naive network inference lags behind the forecast-implied inference (e.g.\ epochs $1005$-$1010$, $1020$-$1030$, $1040$-$1045$, $1085$-$1090$) due to its delayed reaction to update the inferer weights (i.e.\ the naive network inference depends on the historical performance of the infererence workers, see \S\ref{sec:network}). By contrast, the forecasting model identifies the outperforming workers during drift periods, giving them higher weights in the forecast-implied inference.

\subsubsection{Context awareness}
\label{sec:benchmark_context_awareness}
\begin{figure}
    \centering
        \includegraphics[width=0.33\linewidth]{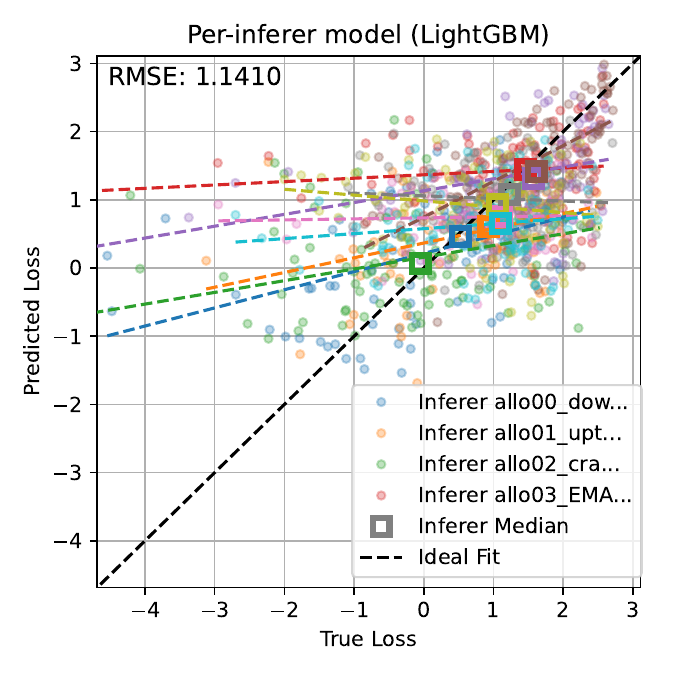}
        \includegraphics[width=0.33\linewidth]{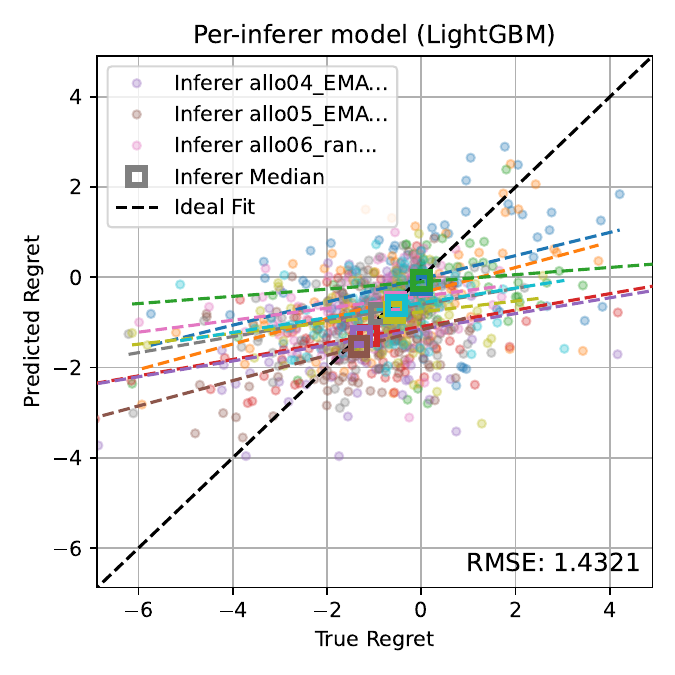}
        \includegraphics[width=0.33\linewidth]{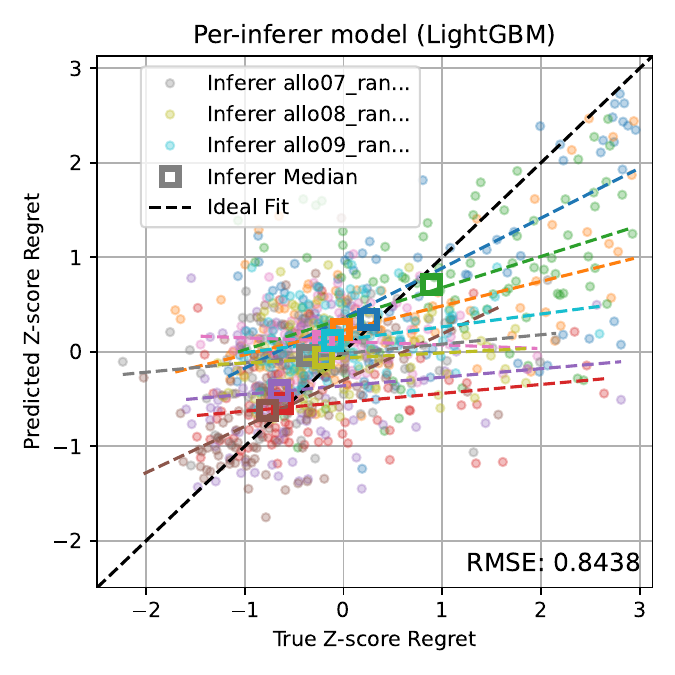}
    \caption{`Context awareness' of forecasting models in the contextual outperformance test. Panels show the true versus predicted properties for forecasters predicting log losses (left), regrets (middle) and regret $z$-scores (right). Point and line styles are as in \autoref{fig:sine_pred}. The legend is split over the three panels; all inferers are present in each panel.}
    \label{fig:context_targets}
\end{figure}
\autoref{fig:context_targets} compares the true and predicted values for default per-inferer forecasting models with the three forecasting target variables: log losses (left panel), regrets (middle panel) and regret $z$-scores (right panel). The forecasters for all targets reasonably predict the medians for each inferer (open squares), such that they approximately follow the ideal one-to-one line in each panel (dashed black line), but the models have varying levels of success in predicting the out- or underperformance of each worker.

The figure highlights the potential benefits or weaknesses of each target variable. For example, although the returns for \textit{allo9} (cyan points) are random and lead to unpredictable losses (left panel, with a linear fit gradient $\approx 0$), converting to regrets enables a limited form of context awareness (middle panel, positive gradient in predictions relative to true values). In turn, the loss model tends to underpredict the losses (i.e.\ overpredict the performance) for the random workers at large true losses ($>1$). However, the underperformance epochs (i.e.\ largest losses during drift periods) for the inferers with EMAs on the true values (red, purple and brown points) become somewhat less predictable with a regret target than with a loss target, resulting in an overall performance decrease ($\mathrm{RMSE}=1.43$) relative to the loss model ($\mathrm{RMSE}=1.14$).

Changing from predicting regrets (middle panel) to the $z$-score of regrets (right panel) rescales the target values such that they become more predictable in epochs of outperformance ($z$-scores $\gtrsim 1$) for the outperforming workers (\textit{allo0}, \textit{allo1} and \textit{allo2}), while retaining the predictable underperformance of inferers with EMAs on the true values (as with a loss target). As with the loss model, the $z$-score model tends to overpredict the performance of the random workers (when the true $z$-score is $<0$). However, the performance gain dominates, making the $z$-score models the most performant models in this experiment ($\mathrm{RMSE}=0.84$).

\subsubsection{Parameter optimization}
\label{sec:benchmark_param_optimization}
\begin{figure}[!t]
    \centering
    \includegraphics[width=0.95\linewidth]{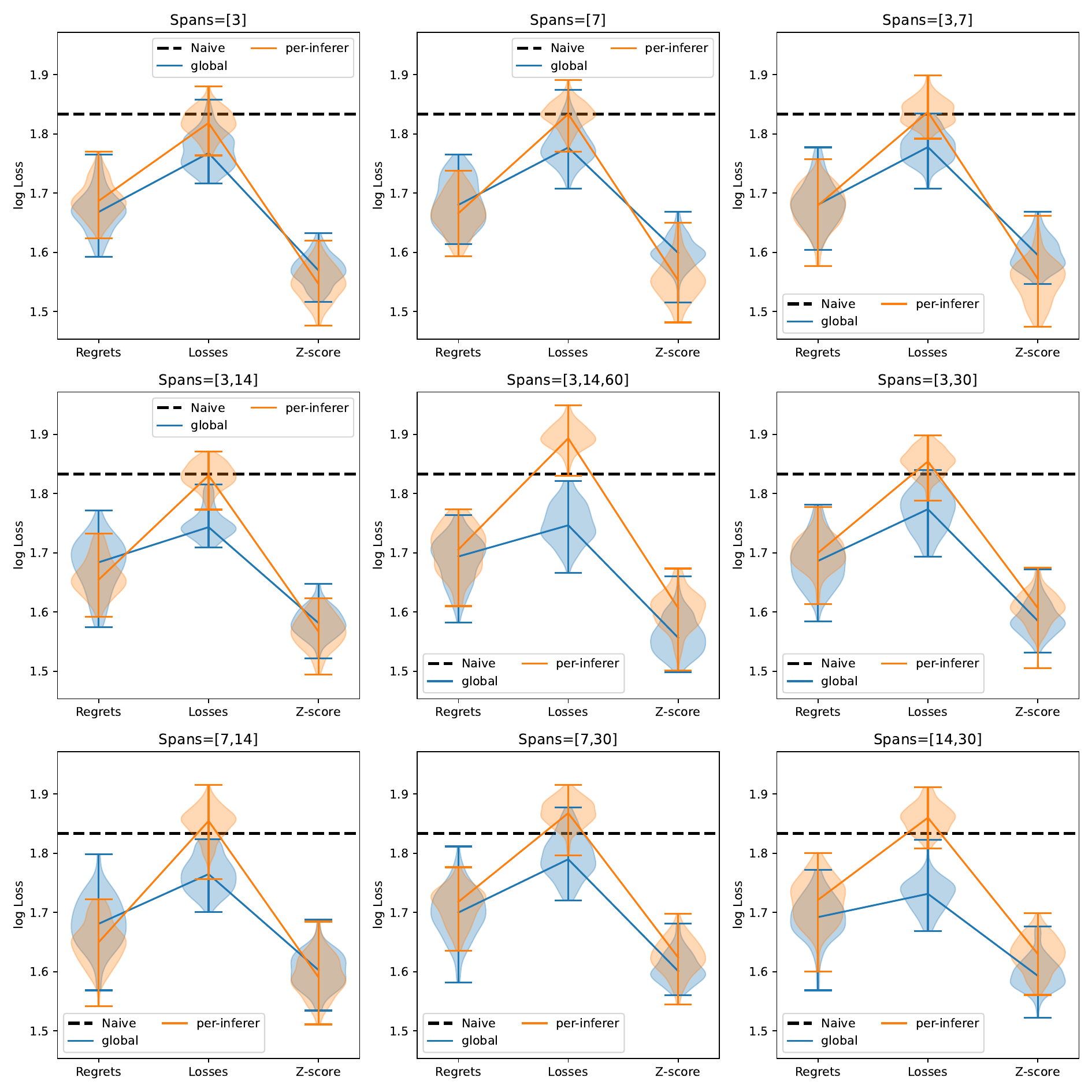}
    \caption{Optimization tests for the LightGBM model (XGBoost shows similar trends) with the contextual outperformance benchmark, with log loss of the (forecast-implied) inferences as the test parameter. Each subpanel shows results for different EMA/rolling property span sets for the feature variables (listed in the subpanel titles), for different forecaster target variables (regrets, losses and regret $z$-scores). The black dashed line shows the naive network inference (i.e.\ without a contribution from forecasting models), while blue and orange lines show the median results for global and per-inferer forecaster models, respectively. Violin plots show the distribution in log loss of 100 repeated tests for each model and feature set combination, to check for variance due to hyperparameter optimization. In these tests, 1000 epochs are used for training and 200 epochs for testing.}
    \label{fig:benchmark_context}
\end{figure}

Due to the variability of forecasts for different seeds in hyperparameter optimization, it is difficult to draw conclusions about the best forecasting model and feature set from single tests. Instead, multiple tests must be performed to compare the distribution in performance metrics. To do so, in \autoref{fig:benchmark_context} we repeat the contextual outperformance tests 100 times for each of the different target variables (loss, regret, $z$-score) and EMA/rolling property spans for the feature variables (9 combinations, indicated in the title of each subpanel). We use the mean log loss of the forecast-implied inference relative to the true values as the property to be minimized. Here, we have increased the number of testing epochs to 200 to reduce the impact of stochasticity in the initial condition generation on the testing period.

The best forecasting model for this test is the per-inferer model predicting $z$-scores with short EMA spans ([3] with log loss $=1.547$, with [7] and [3,7] being the next best span sets with only marginally larger median log losses). For all span sets, $z$-scores are consistently the best performing target variable, followed by regrets and then losses. This indicates that relative performance ($z$-score or regret) provides a simpler target to forecast than absolute performance (losses).

Losses are the only target variable where the global model consistently outperforms the per-inferer model (which often does not outperform the naive inference), potentially as the reduced scatter in predictions in the global model has more impact than the increased context awareness in the per-inferer model for this test. Models with regret as the target (particularly global models) are relatively insensitive to the choice of span set and performance is relatively similar between the global and per-inferer models, though the best regret model is still a per-inferer model (with span set [7,14]). Interestingly, global models with loss and $z$-score targets tend to prefer a combination of longer spans (the best sets are [14,30] for losses and [3,14,60] for $z$-scores) than the per-inferer models (best span set is [3]), highlighting the need for feature sets to be adapted to the particular model architecture.

\section{Experiments with the Allora network}
\label{sec:experiments}
Although controlled experiments are valuable for interpreting test results and pinpointing model weaknesses or areas of improvement, optimizing the forecasting models for real-world performance requires live data. In this section, we repeat the optimization tests from \S\ref{sec:contextual_benchmark}, but instead using data from the Allora testnet from a topic predicting ETH/USD prices in 5 minute intervals. The tests use data obtained between dates 2025-06-02 and 2025-06-23 with six inference workers.

\subsection{Span parameter optimization tests}
\label{sec:exp_param_optimization}
\begin{figure}[!t]
    \centering
    \includegraphics[width=0.95\linewidth]{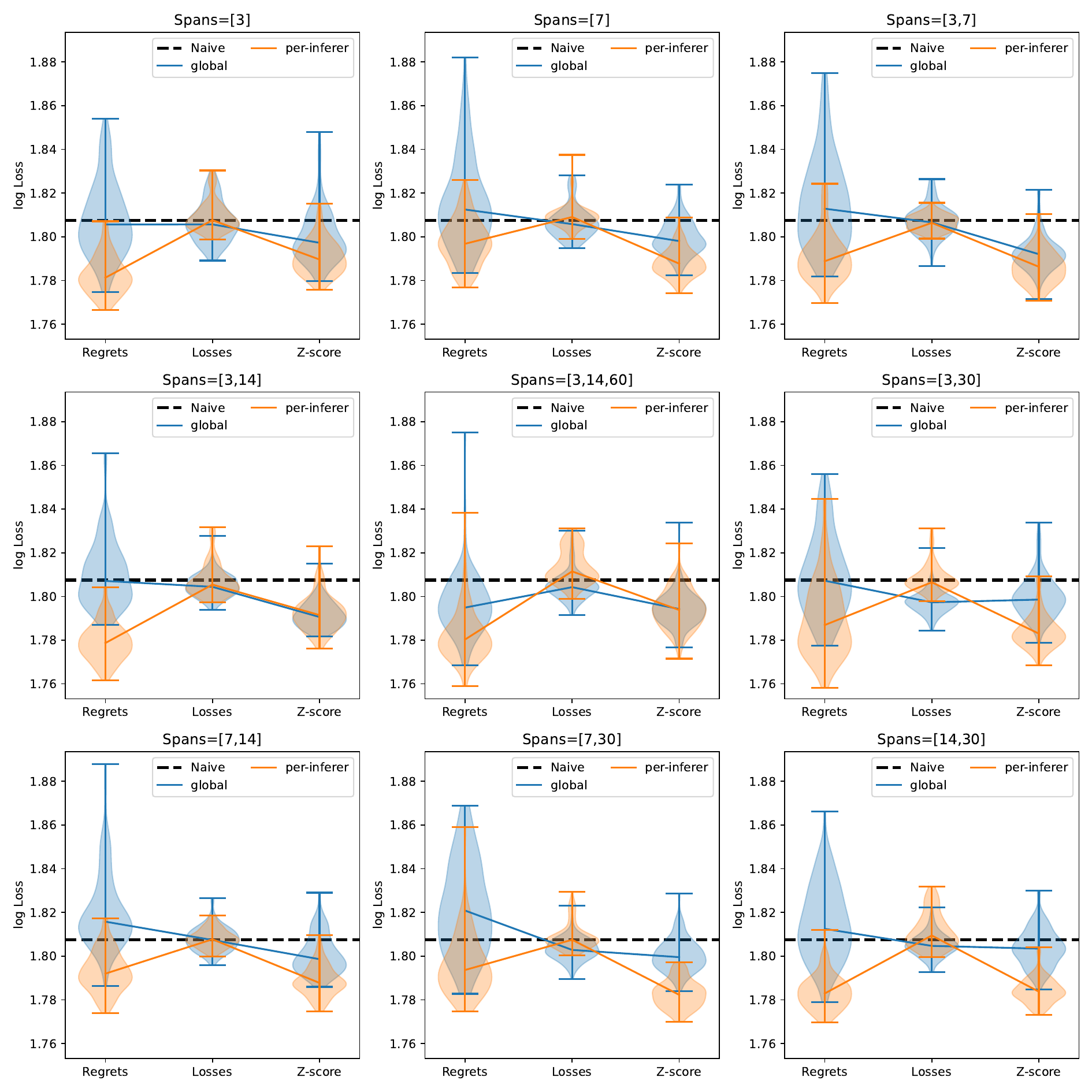}
    \caption{Span parameter optimization tests for LightGBM models with the Allora testnet ETH/USD 5 minute prediction topic. Point and line styles are as in \autoref{fig:benchmark_context}. In these tests, 1000 epochs are used for training and 500 epochs for testing.}
    \label{fig:benchmark_topic13}
\end{figure}
\autoref{fig:benchmark_topic13} repeats the tests from \autoref{fig:benchmark_context} using the live ETH/USD price topic. As before, we use 1000 training epochs for the forecasting models. However, because it is not a controlled experiment, we have increased the testing period to 500 epochs to further reduce sensitivity to the exact testing period.

As expected, improvement of the forecasting models over the naive inference is more modest with live data than in artificial experiments with controlled outperformance periods. With regret as a target, per-inferer models show significant improvement over global models, as only the per-inferer model consistently outperforms the naive inference. All $z$-score models outperform the naive inference in the median, with the per-inferer models generally outperforming the global models (except for span sets [3,14] and [3,14,60], for which both models are similar). Models predicting losses do not consistently outperform the naive network inference, with the global forecasting model still being the preferred setup when losses are used as the target. However, the per-inferer loss models show much closer performance to the global loss models than in the controlled experiments.

In this test, the best overall forecasting model is a per-inferer model with regret as a target and an EMA/rolling property span set of [3,14] (median log loss $= 1.779$). For many span sets (e.g.\ [3,7], [14, 30]), per-inferer regret and $z$-score models often exhibit similar performance. For a loss target, the best span set is [3,30] (median log loss $= 1.797$), while for $z$-scores three span sets ([3,30], [7,30] and [14,30]) have similar levels of performance (mean log losses $=1.782$--$1.783$, within the standard error of the median of $\approx 0.001$). Noticeably, the best performing span sets tend to have the most compact distributions in log losses, i.e.\ they have the least dependence on variations in hyperparameter optimization.

\begin{figure}[!t]
    \centering
    \includegraphics[width=0.95\linewidth]{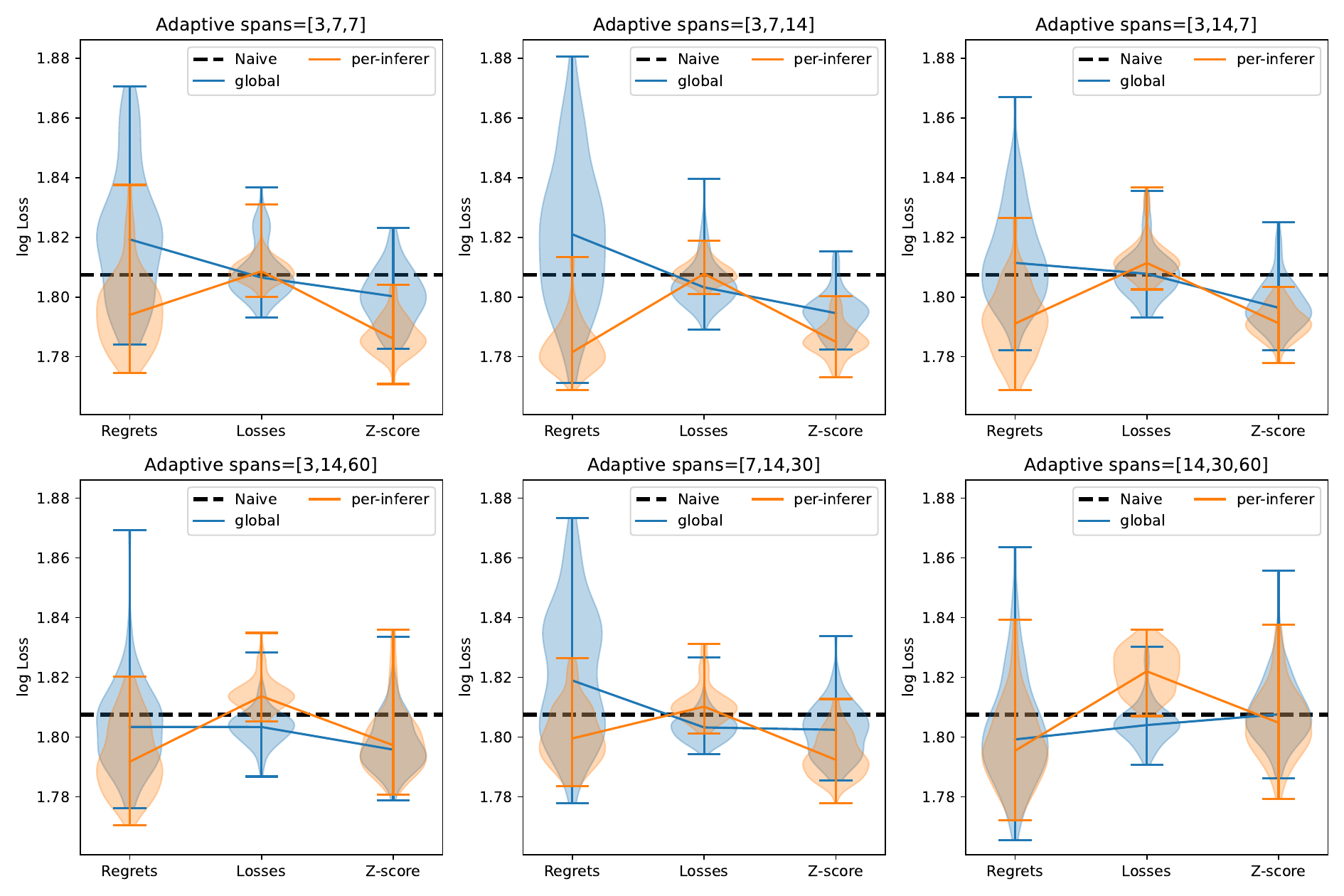}
    \caption{Adaptive span parameter optimization tests for LightGBM model with the ETH/USD 5 minute prediction topic. The adaptive span set use three level for different groups of features, ordered as [gradients, rolling windows, EMAs]. Point and line styles are as in \autoref{fig:benchmark_context}.}
    \label{fig:benchmark_topic13_adaptive}
\end{figure}

In the tests, adding unnecessary features can often make performance worse (e.g.\ for per-inferer models, [3,14,60] has higher median losses than [3,14]).
Therefore, in \autoref{fig:benchmark_topic13_adaptive}, we repeat the optimization tests with an `adaptive' span set that selects three individual spans for three groups of features, ordered as [gradients, rolling windows, EMAs], to test if reduced feature sets can improve performance. For the per-inferer models, we find the adaptive span set [3,7,14] to provide the best performance for both regret (median log loss $= 1.782$) and $z$-score models (median log loss $= 1.785$), although with slightly larger losses (by $\approx 0.003$) than the best models in \autoref{fig:benchmark_topic13}. In the tests, longer span sets (i.e.\ [7,14,30] and [14,30,60]) show reduced performance for regret and $z$-score targets. Longer spans are preferred for EMAs compared to rolling windows, with the span set [3,7,14] outperforming both [3,7,7] and [3,14,7].

\subsection{Training epoch tests}
\label{sec:exp_training_epochs}
\begin{figure}[t]
    \centering
    \includegraphics[width=0.95\linewidth]{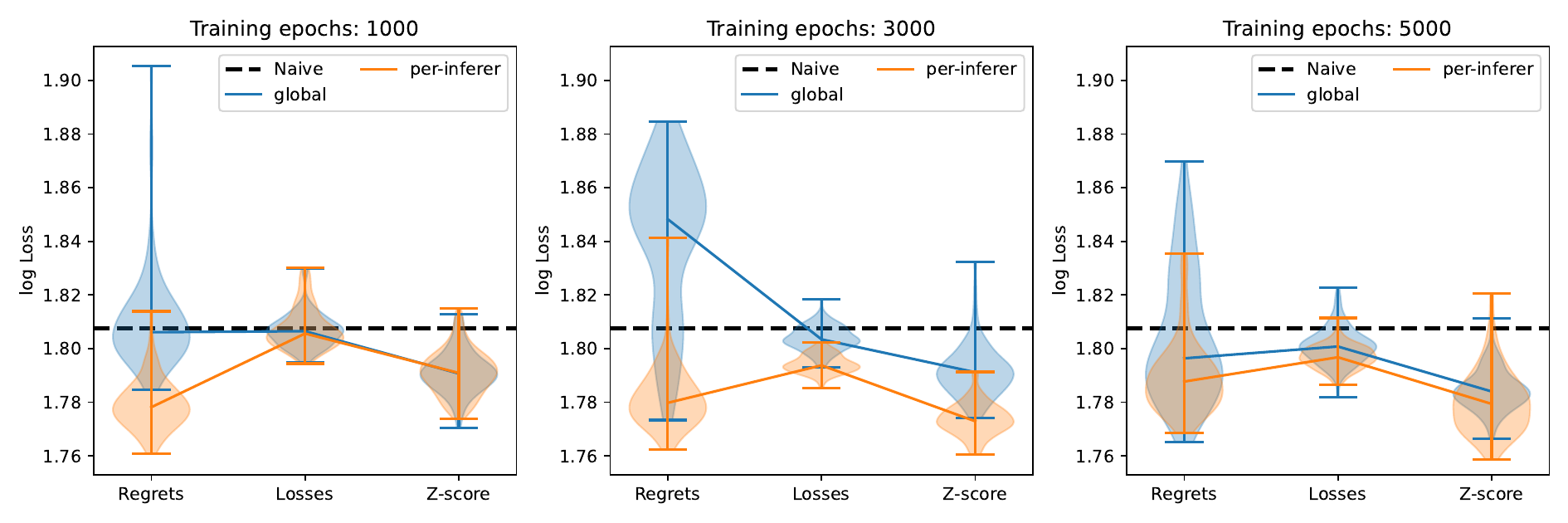}
    \caption{Training epoch number optimization tests for LightGBM model with span set=[3,14] for the ETH/USD 5 minute prediction topic. Point and line styles are as in \autoref{fig:benchmark_context}.}
    \label{fig:Ntrain_3_14}
\end{figure}
\begin{figure}[!t]
    \centering
    \includegraphics[width=0.95\linewidth]{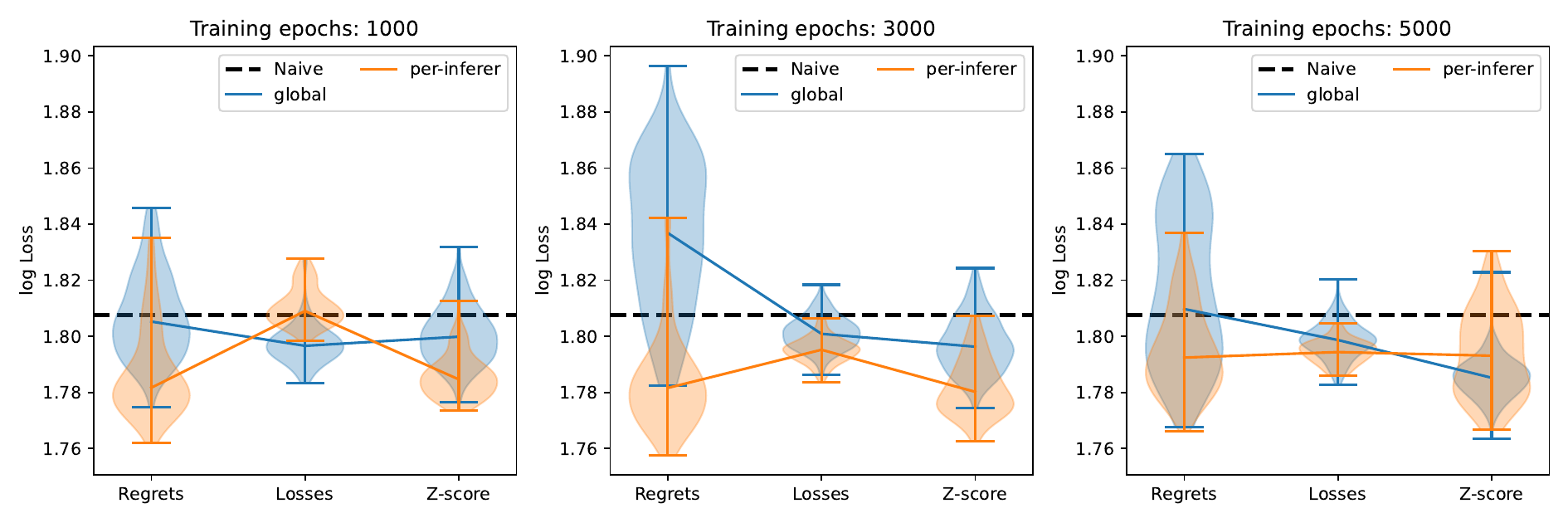}
    \caption{Training epoch number optimization tests for LightGBM model with span set=[3,30] for the ETH/USD 5 minute prediction topic. Point and line styles are as in \autoref{fig:benchmark_context}.}
    \label{fig:Ntrain_3_30}
\end{figure}
\begin{figure}[!t]
    \centering
    \includegraphics[width=0.95\linewidth]{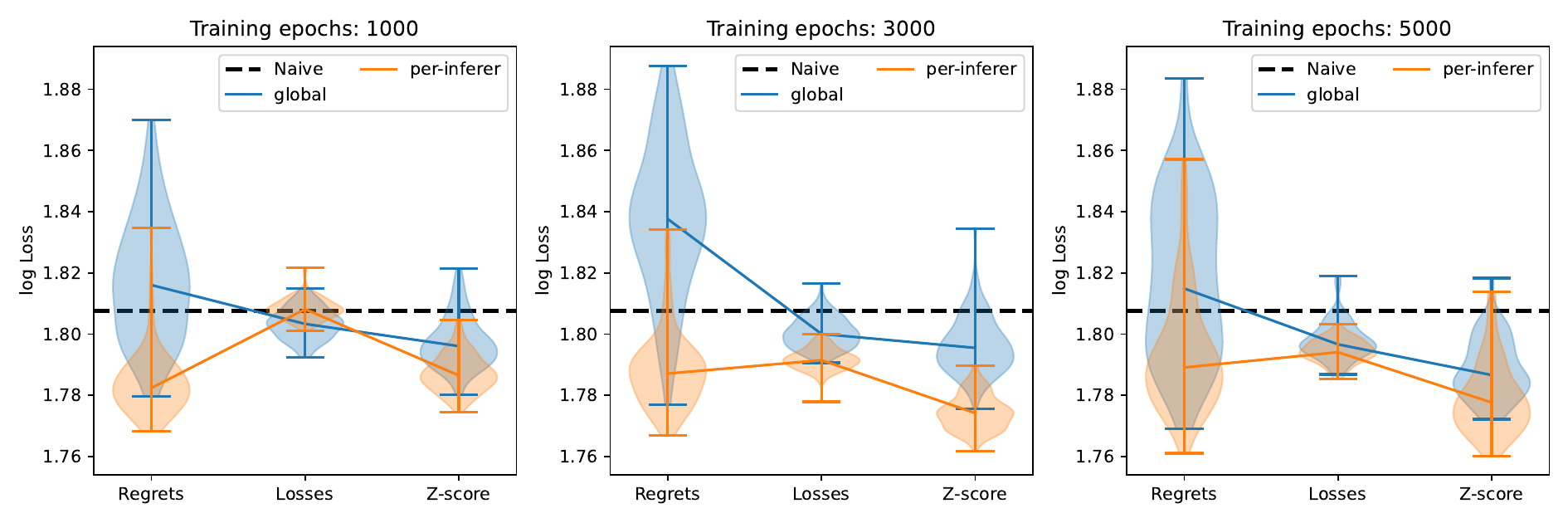}
    \caption{Training epoch number optimization tests for LightGBM model with adaptive span set=[3,7,14] for the ETH/USD 5 minute prediction topic. Point and line styles are as in \autoref{fig:benchmark_context}.}
    \label{fig:Ntrain_3_7_14}
\end{figure}
The previous tests used a fixed number of 1000 epochs for the training period.
Providing more training data can potentially improve model performance, at the expense of increased training times. However, data that are too old may degrade forecaster performance if inferer performance evolves (e.g.\ through retraining) or, in the case of a regret target, the combined network loss has significantly changed over time (thus shifting all regrets). Any improvements from increasing the number of training epochs must therefore be balanced against the requirement to only train on relevant data.

In \autoref{fig:Ntrain_3_14}--\ref{fig:Ntrain_3_7_14}, we retest the best three span sets ([3,14], [3,30] and adaptive [3,7,14], respectively) with increasing numbers of training epochs (1000, 3000, 5000). Interestingly, in all cases the performance of per-inferer regret models degrades as the number of training epochs is increased. For regret targets, the best model still has a span set of [3,14] with 1000 epochs for training. This suggests long term variations (e.g.\ in the combined network loss) may affect the performance of forecasters predicting regrets.

For both loss and $z$-score targets, the best performing models use 3000 training epochs, but in most cases 5000 training epochs also outperforms 1000 training epochs. With increased training epochs (both for 3000 and 5000 epochs), per-inferer models predicting losses can outperform global models and consistently outperform the naive inference, though still fall behind the best regret and $z$-score models. This indicates their inability to compete with the global models in the previous experiments was due to a lack of data. The best performing model with loss as the target property is the adaptive span set [3,7,14] (median log loss $= 1.791$), followed closely by the span set [3,14] (median log loss $= 1.794$). The best performing $z$-score span set becomes [3,14] (median log loss $= 1.773$), making it the best performing model of all tests, with adaptive span set [3,7,14] the next best performing model (median log loss $= 1.774$). Overall, these tests show [3,14] to be the most consistent span set across all target properties, followed by the adaptive span set [3,7,14].

\subsection{Context awareness tests}
\label{sec:exp_context_awareness}
\begin{figure}
    \centering
        \includegraphics[width=0.33\linewidth]{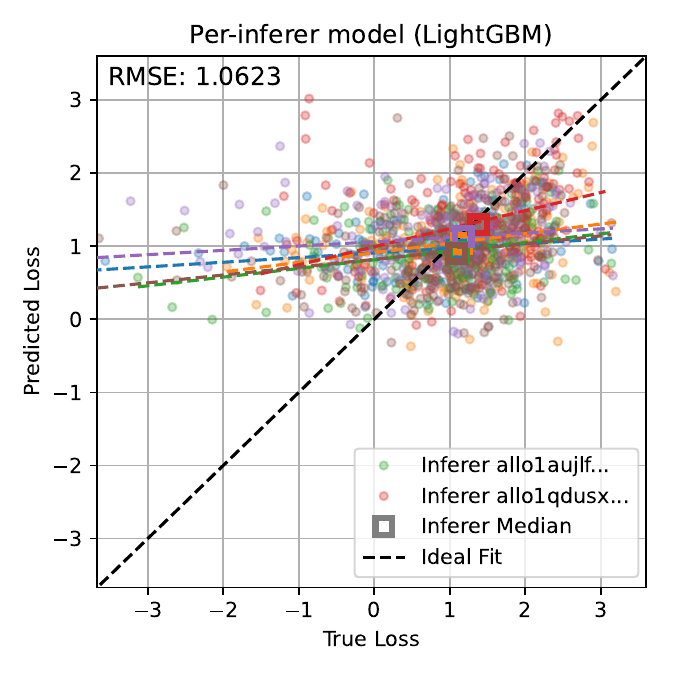}
        \includegraphics[width=0.33\linewidth]{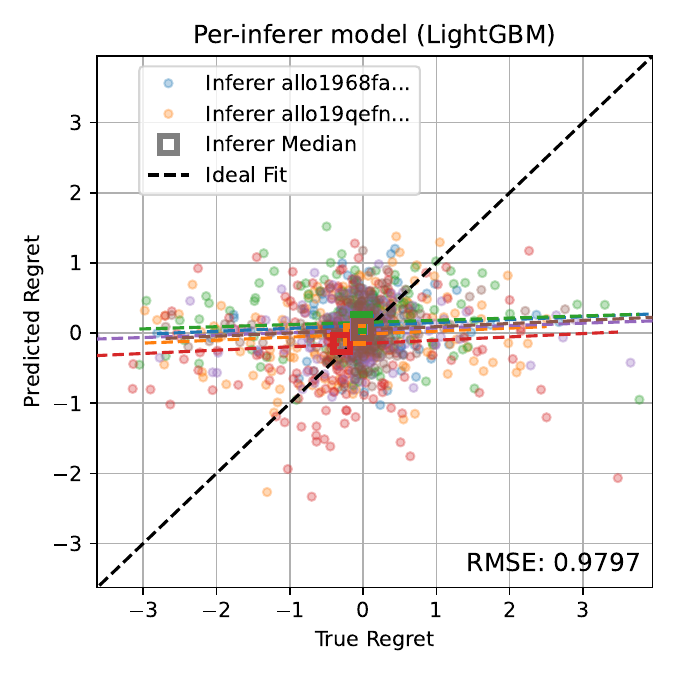}
        \includegraphics[width=0.33\linewidth]{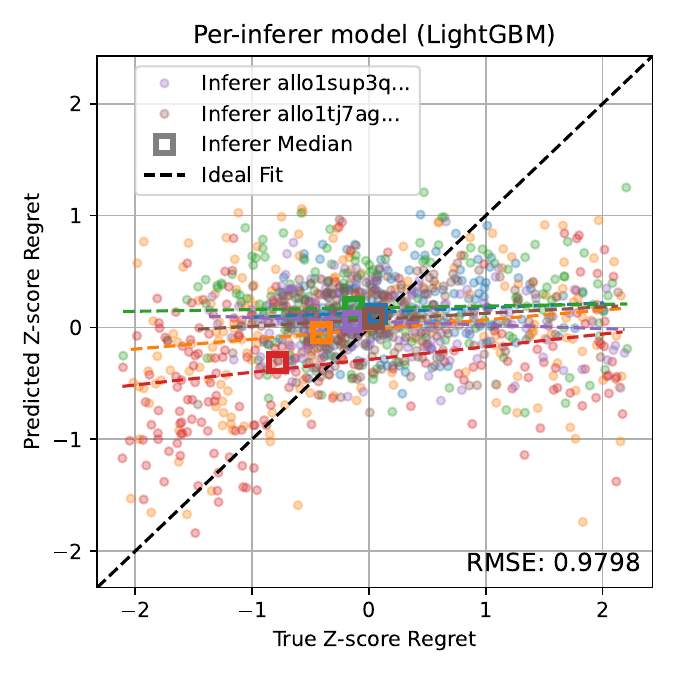}
    \caption{`Context awareness' test for per-inferer forecaster models with the best span set ([3,14]): true versus predicted properties for models forecasting log losses (left panel, 3000 training epochs), regrets (middle panel, 1000 training epochs) and regret $z$-scores (right panel, 3000 training epochs). Point and line styles are as in \autoref{fig:sine_pred}. For clarity, only 200 testing epochs are shown, rather than 500 epochs as in \autoref{fig:benchmark_topic13}--\ref{fig:Ntrain_3_7_14}.}
    \label{fig:topic13_targets}
\end{figure}

In \autoref{fig:topic13_targets} we compare the true and predicted target properties for the best span set ([3,14]) and number of training epochs (1000 for regrets, 3000 for losses and $z$-scores) for the per-inferer models. Relative to the controlled experiments (\S\ref{sec:benchmarks}), context awareness of the forecasting models is more marginal with live network data. The main exception is for the generally underperforming inferer (allo1qdusx\dots, red points), which shows relatively larger gradients in the loss and $z$-score models compared to predictions for other inferers. For most inferers, the linear fits between true and predicted properties show weak but positive gradients that are statistically significant (lower $1\sigma$ confidence intervals on the gradients of $>0$, using bootstrapping of Huber regression).

For the regret target (middle panel), the model identifies many underperformance epochs for inferer allo1qdusx\dots (red points) at true regrets $\sim -1$, but fails to predict the epochs with the lowest regrets ($\lesssim -1.5$). However, these lowest regrets are often due to large temporary increases in the combined network loss, which shifts all inferers to lower regrets than average, making them unpredictable. We note also that the inverse case is not true: large regrets ($\gtrsim 1$) are due to genuine (if temporary) outperformance of inferers, and not due to large changes in the network loss. The models predicting losses and $z$-scores are unaffected by these shifts in the network loss (the former by definition, and the latter because all values are calculated relative to the mean regret). The distribution of true regrets in this test is otherwise very concentrated within $\pm1$, unlike in the contextual outperformance test (\autoref{fig:context_targets}), which may account for the more modest improvement of the forecast-implied inference losses over the naive inference loss in these tests (\autoref{fig:benchmark_topic13}--\ref{fig:Ntrain_3_7_14}).

For the model predicting losses (left panel), the outperformance of inferers is generally not predicted, i.e.\ the predicted loss distribution tends to flatten at true losses $< 1$. This potentially indicates that random chance plays a role in the setting the values of the lowest losses. Instead, the predictive power of losses comes from predicting the underperformance of workers (losses $> 1$), for which there is a general trend of increasing predicted loss with true loss. The mean true losses are similar for most workers (except the one underperforming model), which may explain the lower performance of models with losses as a target compared to regrets and $z$-scores (\autoref{fig:benchmark_topic13}--\ref{fig:Ntrain_3_7_14}).

Relative to the model predicting regrets (middle panel), the $z$-score model (right panel) provides a more even spread in values for the model to predict, particularly by compressing outliers at very large positive or negative regret values. This appears to particularly aid in detecting underperformance ($z$-score $< 1$) for inferers allo1qdusx\dots (red points) and allo19qefn\dots (orange points), but the median predicted $z$-scores for these inferers are well above their median true $z$-scores. In summary, regret and $z$-score models again perform better than loss models.

\section{Discussion and Conclusion}
\label{sec:conclusions}
In this work, we have introduced a model to forecast performance of inferences from participants in a decentralized learning network. We tested a number of model structures (global, per-inferer), target variables (losses, regrets, regret $z$-scores), feature sets (autocorrelation, EMA and rolling mean spans) and training epochs with both synthetic benchmarks and live network data to identify the best performing model(s) for each test. The main findings of this work are as follows.
\begin{enumerate}
    \item Dynamic weighting of predictions that relies only on historical performance is slow to adapt to changes in conditions (e.g.\ the naive inference). So long as the right features can be identified for the machine learning model, performance forecasting models can predict which workers are likely to be more accurate in a given context and weight them higher in the inference combination (\S\ref{sec:contextual_benchmark}). Where the context for performance differences is more difficult to predict, the forecasting model will tend towards predicting the mean performance of each worker (\S\ref{sec:exp_context_awareness}).
    \item Per-inferer models (i.e.\ separate machine learning models for each inference worker) have more `context awareness' than global models (i.e.\ a single machine learning model with inferer ID as a feature) since they isolate the performance of each worker (\S\ref{sec:periodic_benchmark}). In contrast, a global model can effectively stack similarly-performing worker sets to increase training set size and reduce noise.
    \item In both benchmark tests and tests with live data from the Allora testnet (Sections~\ref{sec:benchmark_context_awareness} and \ref{sec:exp_context_awareness}), forecasting models with any of the target variables (losses, regrets, regret $z$-scores) reasonably predict the median performance, but they have varying levels of success in predicting out- and underperformance. In benchmark tests, the $z$-score model is most successful in the context awareness test due to its ability to identify outperformance epochs (\autoref{fig:context_targets}), which is reflected in it being the most accurate model in parameter optimization tests (\S\ref{sec:benchmark_param_optimization}). In experiments with live data, the best-performing models predicting regrets and regret $z$-scores have a similar level of accuracy (\S\ref{sec:experiments}). In practice, it may be beneficial to include multiple forecasting models with different targets such that they can adapt to different situations.
    \item Models predicting losses are the least accurate in both the benchmark and live data tests (\ref{sec:benchmark_param_optimization} and \S\ref{sec:experiments}), despite not being obviously poorer in context awareness tests (\autoref{fig:context_targets} and \autoref{fig:topic13_targets}). This may be due to the weight calculation, which requires the conversion of forecasted losses into approximate regrets using the network loss from the previous epoch (\S\ref{sec:network}). If the network loss is dramatically different from epoch to epoch, then this could shift all regrets into the flat portion of the sigmoid weighting function (\autoref{eq:weight_fn}), making the forecasted losses irrelevant. If losses were instead centered on the mean loss at each epoch, this would effectively obtain the $z$-score target model.
    \item Naturally, feature sets should be optimized to the particular topic and model setup. For example, models with different targets often perform better with different epoch spans for features (\S\ref{sec:benchmark_param_optimization}). Similarly, slight differences in the optimal feature set spans are found between tests with synthetic benchmarks (\S\ref{sec:benchmark_param_optimization}) and live data (\S\ref{sec:exp_param_optimization}).
    \item Increasing training epochs does not necessarily lead to better forecasting model performance (\S\ref{sec:exp_training_epochs}). This could be due to changes in the performance of inference workers over time, for example due to retraining with new data, such that the oldest data becomes stale.
    \item In tests with live data (\S\ref{sec:exp_context_awareness}) the forecasting models (for all targets) struggle to predict outperformance. This might indicate the impact of chance in inferences obtaining the lowest losses. However, \textit{under}performance does often appear predictable, which helps to drive the improvement of the forecast-implied inferences over the naive inference (Sections~\ref{sec:exp_param_optimization} and \ref{sec:exp_training_epochs}).
    \item In some cases, there are large differences in the forecasting model performance depending on differing random seeds for hyperparameter optimization (Sections~\ref{sec:benchmark_param_optimization} and \ref{sec:exp_param_optimization}), although the best performing models often have the most compact distributions of mean log losses. Potentially, ensembling (averaging the same model with different random seeds) could be used for more consistent performance, at the expense of increased computational costs.
\end{enumerate}

In developing a performance forecasting model for decentralized learning networks, this work builds on a significant body of literature on model combination and aggregation \citep[e.g.\ see the recent review by][]{Wang_et_al_2023}. However, it differs in a number of fundamental respects compared to previous studies. Our forecasting model predicts performance of inferences, which are then converted to combination weights, rather than predicting weights directly. Most studies using machine learning to predict combination weights are aiming to optimize constant weights over a full time series \citep[e.g.][]{Prudencio_and_Ludermir_06, Lemke_and_Gabrys_10, Montero-Manso_et_al_20, Kang_et_al_22}. Here, the model makes predictions at every epoch in a time series such that the model can be context aware. As combination weights are not a natural output of the learning network, they would need to be calculated independently at each epoch. However, for a single epoch, such weights will be non-unique in all but the simplest cases. Additionally, the model must account for differing inferer sets at different epochs. Even if ideal weights could be identified at each epoch, these may not be relevant if the inferer set changes, so the weight and model would need to be retrained each time (e.g.\ consider a set where two of the workers always have predictions on opposite sides of the true value, and one of the workers suddenly drops out of the set). Although there exist models that combine multiple predictions at different epochs with machine learning, this is achieved by predicting the combined value itself, rather than weights \citep[e.g.\ using neural networks,][]{Zhao_and_Feng_20}. In the case of forecasting workers for decentralized learning networks, weights are also required for scoring and reward allocation \citep{K24}.

This work is, of course, not intended to be a comprehensive performance forecasting model for all topics and tasks. Rather, it is a demonstration of dynamic performance forecasting and a starting point for further development, and provides lessons and potential improvements for future work. As is generally the case in machine learning, feature choice and engineering is a crucial aspect in enabling well-performing models. Care should be taken to identify important features and optimize them for the particular topic and model set. Although the performance forecasting model was designed and tested for use in a decentralized learning network, dynamic performance forecasting is applicable in any combination problem where the most accurate models change on short timescales.

\section*{Data Availability and Acknowledgments}
The standard performance forecaster model for the Allora network is available  in the network's Github repository at \href{https://github.com/allora-network/allora-forecaster}{https://github.com/allora-network/allora-forecaster}. COOL Research DAO is a Decentralised Autonomous Organisation supporting research in astrophysics aimed at uncovering our cosmic origins \citep{chevance25}.

\begin{sloppypar}
\bibliographystyle{ADI}
{\small
\bibliography{bibliography}
}
\end{sloppypar}

\end{document}